\def\year{2023}\relax
\documentclass[letterpaper]{article} 
\usepackage{aaai22}  
\usepackage{times}  
\usepackage{helvet}  
\usepackage{courier}  
\usepackage[hyphens]{url}  
\usepackage{graphicx} 
\usepackage{natbib}  
\usepackage{caption} 
\DeclareCaptionStyle{ruled}{labelfont=normalfont,labelsep=colon,strut=off} 
\usepackage{algorithm}
\usepackage{algorithmic}

\usepackage{booktabs}
\usepackage{amsmath}
\usepackage{amssymb}
\usepackage{svg}

\usepackage{newfloat}
\usepackage{listings}
\floatstyle{ruled}
\newfloat{listing}{tb}{lst}{}
\floatname{listing}{Listing}
\usepackage{soul}

\title{It's All Relative: Interpretable Models for Scoring Bias in Documents}
\author {
    Aswin Suresh,\textsuperscript{\rm 1}
    Chi-Hsuan Wu, \textsuperscript{\rm 2}
    Matthias Grossglauser \textsuperscript{\rm 1}
}
\affiliations {
    \textsuperscript{\rm 1} EPFL, Switzerland\\
    \textsuperscript{\rm 2} HKUST, Hong Kong\\
    aswin.suresh@epfl.ch, cwuau@connect.ust.hk, matthias.grossglauser@epfl.ch
}

\usepackage{bibentry}

\begin{document}
\nocopyright
\maketitle

\begin{abstract}
We propose an interpretable model to score the bias present in web documents, based only on their textual content. 
Our model incorporates assumptions reminiscent of the Bradley-Terry axioms and is trained on pairs of revisions of the same Wikipedia article, where one version is more biased than the other.
While prior approaches based on absolute bias classification have struggled to obtain a high accuracy for the task, we are able to develop a useful model for scoring bias by learning to perform pairwise comparisons of bias accurately.
We show that we can interpret the parameters of the trained model to discover the words most indicative of bias. 
We also apply our model in three different settings - studying the temporal evolution of bias in Wikipedia articles, comparing news sources based on bias, and scoring bias in law amendments. 
In each case,  we demonstrate that the outputs of the model can be explained and validated, even for the two domains that are outside the training-data domain. We also use the model to compare the general level of bias between domains, where we see that legal texts are the least biased and news media are the most biased, with Wikipedia articles in between. Given its high performance, simplicity, interpretability, and wide applicability, we hope the model will be useful for a large community, including Wikipedia and news editors, political and social scientists, and the general public.
\end{abstract}

\section{Introduction}
\label{sec:intro}

\noindent In recent years, the amount of human-generated data on the Web has grown exponentially in size and relevance.
Some prominent examples include encyclopedias (such as Wikipedia), online news portals, and political data (such as legislative acts and speeches).
As they are generated by humans, these data suffer from bias to varying degrees, which results from the specific worldview and agenda of their authors.

Here we use the term bias to mean \textit{inappropriate subjectivity}, as defined in \cite{pryzant2020wnc} - ``Subjective bias occurs when language that should be neutral and fair is skewed by feeling, opinion, or taste (whether consciously or unconsciously)". 
Data affected by such subjective bias inform the perspectives and influence the decisions, both political and otherwise, of an increasing number of people.
When people are unaware of the bias present in the data, their ability to arrive at accurate conclusions is compromised.
This lack of awareness also contributes to the formation of echo chambers and makes it difficult to build consensus for actions towards the common good.
Therefore, it is important to identify and measure this bias and to do so in an explainable manner so as to be trustworthy.

Currently, this is done manually in several domains: Wikipedia editors mark articles and edits as violating neutrality, companies such as AllSides \cite{allsides2022allsides} provide ratings of bias in the media, and political scientists analyze speeches to study subjective language as expressions of ideological positions.
However, such manual analysis cannot scale to the exponentially growing size of web data, hence necessitating the use of automated approaches.
Machine-learning models that can benefit from the large training data are of particular interest in this regard.

The English Wikipedia is in many ways an ideal source of training data for these models.
It has a Neutral Point of View (NPOV) policy \cite{wikipedia2022npov}, the adherence to which can be used as a measure of unbiasedness (neutrality).
The policy requires following principles such as not stating opinions as facts (and vice versa), not using language that sympathizes with or disparages the subject etc.
Wikipedia also has an active community of editors that enforces this policy by making edits to reword or remove problematic content from articles and leaving comments to indicate NPOV issues.
Moreover, the data is extensive due to Wikipedia's vast collection of articles spanning a wide range of subjects, and the complete revision history of these articles, along with the editors' comments, is accessible to the public.

However, it is worth noting that using Wikipedia's NPOV policy and the actions of editors to enforce the policy as standards of neutrality are not without criticism.
\cite{matei2011wikipedia} show that NPOV policy itself and its interpretation is subject to much conflict between Wikipedia editors.
\cite{keegan2017evolution} study the evolution of rules in Wikipedia and note that `rules-in-form' such as the NPOV policy may still be subject to deliberation and revision for a long period of time.
In a way, this reflects the underlying difficulty of agreeing on a definition of what is subjective; in other words subjectivity itself seems to be subjective.
This problem is mitigated to some extent if we consider comparisons of subjectivity, which seem to be more objective as we also discuss in the following paragraphs.
In addition, making the reasoning behind a bias decision by the model transparent allows users to re-assess it if needed so as to reduce the influence of subjective standards in the decision.
 
On the other hand, an alternate view of the NPOV revision dynamics is provided by \cite{pavalanathan2018mind}, where it is seen that the NPOV policy encourages articles to converge to a common standard of neutrality, as judged by several lexicons, in spite of the editors themselves not changing their styles.
This suggests that in many cases editors allow edits by others that reduce bias in terms of the community's interpretation of the NPOV policy, even if they themselves would not make such edits or consider them necessary.
We did a manual analysis of 100 random POV-related edits (identified automatically based on editor comments). We agreed that 82 of them reduced bias, and while we didn't think the other 18 necessarily reduced bias, we did not consider them to be increasing bias or to be significantly harmful to the article. 
Therefore, overall we think the approach of training a model on POV-related edits to quantify bias (with regard to the Wikipedia community's definition and interpretation of neutrality) is reasonable given that the users of the model keep in mind its limitations, which we again discuss at the end of the paper under ethical considerations.

Our goal in this work is to develop such a model trained on POV-related edits to Wikipedia articles that can quantify bias in web documents and study its applicability to Wikipedia itself, as well as to two domains outside the training data, namely news and legal texts. 
Importantly, in addition to being reasonably accurate, we also want the model to be \textit{interpretable}, i.e., we would like to use the parameters of the trained model to infer the words indicative of bias and to explain the output of the models.
Interpretability is useful to gain an understanding of how bias occurs and to make the model more trustworthy and easy to debug for users. 






Previous work on bias modelling  predominantly considers the task of bias classification of short pieces of texts (e.g. words and sentences) in an \textit{absolute} sense, i.e., classifying a given piece of text as biased or unbiased. \cite{pryzant2020wnc} \cite{zhong2021wikibias} \cite{li2022towards}.
However, we suggest that classifying general web documents in this manner is not a well-defined task for two reasons.

First, the threshold for deciding whether a text is biased or not is subjective, especially for longer texts such as documents. 
In fact, previous work has found poor inter-annotator agreement when obtaining ground-truth labels \cite{lim-etal-2020-annotating}\cite{de-kock-vlachos-2022-leveraging}\cite{SPINDE2021102505}.

Second, this threshold could vary depending on the topic and the domain of the document.
For instance, a Wikipedia article considered `unbiased' on a politically controversial topic is arguably prone to have more subjective statements than a `biased' one describing an objective scientific truth.
A `biased' Wikipedia article can still be more objective than a relatively `unbiased' political speech.

Hence, in this work, we instead consider the task of \textit{relative} bias classification of documents.
We define this as the task of predicting which text, among a given pair of texts, is more biased.
Solving such a task does not require the determination of a bias threshold and therefore avoids the above two issues.
Previous works have found greater inter-annotator agreement and higher human accuracy for this task, compared to absolute-bias classification \cite{de-kock-vlachos-2022-leveraging} \cite{aroyo2019crowd}.

In addition, we can get abundant training data for this task from the revision history of Wikipedia articles.
Each time a Wikipedia editor corrects a POV issue present in an article version, a pair of texts is generated where one text (the version before the correction) is more biased than the other (the version after the correction).
Comparing bias at the document level instead of the word or sentence level also allows us to benefit from additional context information such as the overall topic of the document.

Solving this task can enable us to gain insights into the features indicative of bias and can possibly provide a mechanism for quantifying bias in documents across time, topics, and domains.

Within Wikipedia, such models can be used to monitor the evolution of bias in articles as they are revised over time.
This can help to draw the attention of editors toward biased articles and revisions. 
It would also be useful to study the speed of bias mitigation in peer-produced texts for researchers in this field.

News, political speeches and legal texts are some other domains where bias detection is important, but unlike Wikipedia, data annotated for bias is scarce in those domains which makes training models difficult.
If the models trained on Wikipedia can generalise to these other domains, it can address this problem to some extent.

We seek to answer the following research questions through this work:
\begin{itemize}
    \item \textbf{RQ1:} Given a pair of consecutive revisions (versions) of the same Wikipedia article that are generated when a POV issue is corrected, how well can we predict which among them is more biased (i.e., the version before the correction), using only models based on their \textit{textual} content?
    \item \textbf{RQ2:} Can we understand which words are correlated with bias by interpreting the parameters of the predictive models?
    \item \textbf{RQ3:} How widely can such predictive models generalise?
    Can the models that are trained only to compare consecutive revisions of the same article generalise across time, topic and domain?
    \begin{itemize}
        \item \textbf{RQ3a:} Can the models compare revisions of the article that are not consecutive and thereby capture the temporal evolution of bias of an article over its entire history? 
        \item \textbf{RQ3b:} Can we compare bias in different Wikipedia articles from different topics using the predictive models?
        \item \textbf{RQ3c:} Can the predictive models, which are trained only on Wikipedia, generalise to other domains of text? Can we use them to compare the level of bias between different domains?
    \end{itemize}

\end{itemize}

We make the following contributions:

\begin{itemize}
    \item Towards answering RQ1, we develop discrete choice models for relative bias classification using only the article's textual content as features. We compare the accuracy of our model against both random chance and strong baselines.
    \item We use the parameters of the trained models to compute a bias score for words that gives the contribution of the word to a document being biased. We use the scores to discover words that are indicative of bias (RQ2).
    \item We also use the trained models to compute a bias score for documents. Although the models are trained only to compare consecutive revisions of Wikipedia articles, the score that we compute using it can be used to compare bias in non-consecutive revisions of an article, different Wikipedia articles, and even documents from different domains including laws and news articles. We analyse the computed scores to answer RQ3.
    \item We curate new datasets of Wikipedia articles to train and evaluate our models. We will release all the datasets and our code publicly when the paper is published.
\end{itemize}
    
The paper is structured as follows.
Section \ref{sec:relwork} describes the related work.
Section \ref{sec:dataset} provides details about datasets we use and have curated for this study.
Section \ref{sec:model} describes the bias model in detail.
In Section \ref{sec:eval}, we evaluate the performance of the model, explore its interpretability, and comment on some potential applications of the model to other domains.
We conclude the paper in Section \ref{sec:conclusion}.

\section{Related Work}
\label{sec:relwork}

Bias in web documents is a well-studied topic.
Previous works have studied subjective bias in Wikipedia \cite{wong2021wiki} \cite{pryzant2020wnc} \cite{zhong2021wikibias} \cite{de-kock-vlachos-2022-leveraging}, news media \cite{lim-etal-2020-annotating} and political speeches \cite{vafa-etal-2020-text}.

In \cite{wong2021wiki}, the authors collect a dataset of pairs of biased and neutral versions of the same Wikipedia article. 
They do this by going through the revision history of articles and picking revisions where the POV template (which is used to flag NPOV issues) was added by editors where it was removed.
However, the dataset size is quite small (only about 5,000 pairs) as the template is not updated often.
The authors only use metadata to develop classification models to predict if a given article version is biased or not, and they achieve an accuracy only slightly better than random (52\%).

In \cite{pryzant2020wnc} and \cite{zhong2021wikibias}, the authors work on the task of identifying bias in words and sentences.
They obtain a parallel corpus of about 200,000 pairs of biased and neutral sentences by aligning the sentences before and after an NPOV-related revision using the BLEU score.
Such revisions are identified by checking for the presence of regular expressions denoting NPOV issues in the editor's comments. 
They use this corpus to train models for the binary classification task of predicting whether a word or sentence is biased.
For the task of sentence bias prediction, the best model in \cite{zhong2021wikibias} achieves an accuracy of 73\%, after additional fine-tuning on a manually annotated set of sentence pairs.
\cite{li2022towards} consider the problem of bias detection when annotations are scarce, noisy and potentially biased. 
By using data augmentation and a self-supervised contrastive learning objective, they are able to achieve a similar performance as \cite{zhong2021wikibias} with a much smaller dataset.
None of these works considers the task of predicting bias at the document level. 

In \cite{de-kock-vlachos-2022-leveraging}, the authors consider the related task of promotional tone detection at the document level.
Like \cite{wong2021wiki}, they also use template information to collect the dataset and hence have a relatively small dataset size.
They use neural network models with a gradient reversal layer to prevent the model from learning features that are topic specific to enable better generalisation.
Their best model achieves an accuracy of 64.3\% for predicting if a given text has a promotional tone.


The models above are based on deep neural networks and therefore require significant time and GPU resources for training and inference.
In particular for training, the models in \cite{pryzant2020wnc} and \cite{zhong2021wikibias} need several hours while the model in \cite{de-kock-vlachos-2022-leveraging} needs more than a day.

The models are also complex and hence difficult to interpret.
Although an explanation can be given for which parts of a \textit{given} text are biased, it is difficult to answer which words \textit{in general} are indicative of bias based on the trained model.

Moreover, all the models above are trained for the task of bias prediction in an absolute sense while our model is trained for the pairwise comparison of bias, which is a better-defined task as described in Section \ref{sec:intro} \footnote{The authors in \cite{de-kock-vlachos-2022-leveraging} additionally test their model for ranked prediction of promotional tone and report an accuracy of 74.1\% but the model is not trained for this task.}.

Our models are instances of the class of discrete-choice models that are regularly used for several applications involving learning from pairwise comparisons. 
Such models have been used for predicting the survival of edits in Wikipedia without using the edit text \cite{yardim2018wikipedia}, predicting the outcome of football matches \cite{maystre2019pairwise} and  predicting the success of amendments in the European Parliament \cite{kristof2021war}.
In the latter, the authors also study the use of text features in combination with other features.

To the best of our knowledge, we are the first to propose modelling bias in documents using a framework of pairwise comparisons.
We do so using a discrete-choice model based only on the document text, that is easily interpretable and relatively inexpensive computationally to train and use compared to prior bias models, while also achieving similar or better accuracies.
We also study the application of the model to a variety of document domains and demonstrate its generalisability and the insights that could be gathered from it.

\section{Datasets}
\label{sec:dataset}
We use four datasets in this paper, two of which we collected ourselves.
We present a summary of the statistics of each dataset in Table \ref{tab:datstat}.
We will now briefly describe the datasets.

\subsection{Wikipedia: Article Neutrality}

To train and evaluate our model, we collect a new dataset that we call the Wikipedia Article Neutrality Dataset (WAND). 
The dataset can be viewed as an article-level version of the sentence pair dataset collected in \cite{zhong2021wikibias}.

The dataset consists of the text of pairs of revisions of the same Wikipedia article where one revision is more biased than the other.
We collect it by going through the revision history of all articles in the English Wikipedia and collecting a pair of revisions before and after a POV-related edit is made.
We identify POV-related edits by checking for the presence of certain regular expressions; we use the same list of expressions used in \cite{zhong2021wikibias}.

For each revision, we use the \texttt{mwparserfromhell} package \cite{kurtovic2022mwparser} to parse its \textit{wikitext} as obtained from the MediaWiki API.
We then apply the text pre-processing steps, followed in \cite{wong2021wiki} and \cite{pryzant2020wnc}, to keep only the plain text (excluding wikilinks, templates, and tags) from the main content part of the article (excluding the External Links and References sections).

For every pair, we compute the Levenshtein edit distance between the texts of the revisions and keep only the pairs that have a distance of at least ten.
We do this to remove pairs where only minor edits such as corrections in spelling and punctuation were made. 

\subsection{Wikipedia: Controversial Issues}
\label{sec:wikicontro}
As the WAND dataset contains the revisions at only the times of the POV-related edits, we cannot use it to evaluate the performance of our models in estimating bias changes over the whole history of the articles.
Therefore, we construct a new dataset of revisions of the articles mentioned in Wikipedia's \textit{List of Controversial Issues} \cite{wiki2022clist}.
Wikipedia editors are urged to regularly check these articles to make sure that the presentation follows the NPOV policy, as they are frequently subjected to biased edits.
The articles in the list are organised by topics, such as \textit{Politics and Economics}, \textit{History}, \textit{Religion}, etc.

For each article, we collect the text for 100 revisions periodically sampled from its history.
The number of revisions $k$ between each sampled revision is different for each article as some articles have more revisions in it history than others (depending on age or frequency of editing of the article).
The text is pre-processed, as in WAND, to retain only the plain text from the main article content.
To assign POV and NPOV labels to the revisions, we make use of comments attached to the $k$ edits that were made immediately before and after each revision.
We identify POV related comments by using the same regular expressions used when constructing WAND. 
For each revision $r$, we then compute the proportion of POV comments $\widehat{c_r} = \frac{c_r}{2k}$.
We label revisions that are in the upper quartile of $\widehat{c_r}$ for the whole corpus as POV and revisions that are in the lower quartile as NPOV.  

\subsection{News}
To study how well models of bias generalise to domains that are different from their training domain we apply our models trained on Wikipedia to score bias in news articles.
We use the Webis Bias Flipper-18 dataset \cite{chen2018bias} that contains news articles from outlets with different ideological biases (left, right and center).
The articles that describe the same event are grouped into stories, which enables us to eliminate the effect of the event itself by ranking articles within each group.

The grouping of the news articles and the ideological bias labels of the outlets come from AllSides.com \cite{allsides2022allsides}.
This website aims to present balanced coverage of news by presenting articles from outlets with different ideological biases.
The ideological bias labels for each outlet are determined by a combination of factors, including editorial review and community feedback. 

\subsection{European Parliament: Law Amendments} 
Texts in the legal domain are likely to have much less subjective language relative to general texts. 
To see if the model can score these correctly, we evaluate it on a dataset of law amendments.
We use the dataset of amendments proposed in the 8th term of the European Parliament, released in \cite{kristof2021war}.

Each amendment in the dataset consists of a pair of texts.
The first text is a paragraph of the law in the original law text as drafted by the European Commission, which is the executive branch of the European Union and the body in charge of drafting new laws.
The second text is the amended version of the same paragraph as proposed by a parliamentarian or a group of parliamentarians when the law is being discussed within the committees of the European Parliament.
An amendment consists of multiple edits, where an edit is a contiguous block of text that is deleted, inserted or replaced. 

Proposed amendments are voted on within the committee and a subset of edits in them are accepted for incorporation into a modified draft law which is subsequently presented at a plenary meeting of the parliament.
For each edit, we know if it was accepted for incorporation into the modified draft or not.

\begin{table}[h]
\centering
\begin{tabular}{@{}lr@{}}
\multicolumn{2}{c}{\textbf{Wikipedia Article Neutrality}}               \\ \midrule
Number of articles                        & 358,941               \\
Number of revision pairs                        & 895,957               \\
Median revision pairs per article               & 1               \\
& \\
\multicolumn{2}{c}{\textbf{Wikipedia Controversial Issues}} \\ \midrule
Number of articles                        & 1,544               \\
Median history length                     & 4,729               \\
& \\
\multicolumn{2}{c}{\textbf{News}}              \\ \midrule
Number of stories                         & 2,781               \\
Number of outlets                         & 77               \\
Number of articles                        & 6,448              \\
& \\
\multicolumn{2}{c}{\textbf{European Parliament Amendments}}              \\ \midrule
Number of original texts                         & 28,407               \\
Number of proposed amendments                        & 98,245               \\
Amendments with at least 1 accepted edit                       & 37,689  \\
Amendments with at least 1 rejected edit                       & 73,604 \\
\end{tabular}
\caption{Dataset statistics}
\label{tab:datstat}
\end{table}

\section{Model}
\label{sec:model}
We now describe the model we propose for ranking documents by bias.
Model interpretability is one of our primary concerns. 
Hence we design the whole pipeline from feature extraction to prediction while taking this into account.
In particular, we generally avoid using multi-layer neural networks, except in some models where we use it for feature extraction at the word level.

\subsection{Features}
To represent the text of a document, we use the normalised sum of the embedding vectors of the words in the text.
We use two kinds of word embeddings, namely static and contextual.

Static word embeddings represent the meaning of each word by a single vector in $d$-dimensional space.
If a word has different meanings in different contexts, the single vector represents a weighted average of the different meanings based on their frequency.
The embedding is obtained by training each word's embedding vector so as to predict the words that appear in its context, given their embedding vectors. 
We use pre-trained Fasttext word-embeddings \cite{bojanowski2017enriching} that were trained on the English Wikipedia.

Contextual word embeddings, on the other hand, represent the meaning of a word in the context where it appears, hence each \textit{occurrence} of the word (a \textit{token}) is represented by a single vector.
The BERT model \cite{devlin2019bert} is arguably one of the most commonly used contextual embeddings and has been used in prior work in bias modelling at the sentence level \cite{zhong2021wikibias}. However it can only model sequences up to a maximum length of 512 tokens due to the quadratic complexity of the attention mechanism, and thus cannot effectively model long documents such as Wikipedia articles. 

We, therefore, use a pre-trained Longformer model \cite{Beltagy2020Longformer}, which is a variation of BERT that uses sliding window attention, enabling it to model long sequences efficiently.
It has been used to model Wikipedia articles in prior work \cite{de-kock-vlachos-2022-leveraging}.
Specifically, we use the longformer-base-4096 model from HuggingFace \cite{allenai2022longformer}.

When using static embeddings we obtain the vector representation of a text $i$ as
\begin{equation}
    \label{eq:ahat}
    \mathbf{\widehat{t_i}} = \frac{\mathbf{t_i}}{\lVert \mathbf{t_i} \rVert },
\end{equation}
where
\begin{equation}
    \label{eq:t}
    \mathbf{t_i} = \sum_{w \in \mathcal{V}_i} n_i(w) \mathbf{v_w}.
\end{equation}
Here $\mathcal{V}_i$ is the set of words in text $i$, $n_i(w)$ is the frequency of word $w$ in text $i$ and $\mathbf{v_w}$ is the embedding vector of the word.
The representation $\mathbf{t_i}$ is obtained in a similar manner when using contextual embeddings except that we consider tokens instead of words.

We convert text to lowercase before computing the vector representation for the models as we find that this gives better performance on the validation set.

\subsection{Model Architecture}
Our model takes inputs in the form of \textit{pairs} of texts and predicts which is more biased than the other.
We use a model reminiscent of the Bradley-Terry model of pairwise comparison outcomes \cite{bradley1952comparison}.

We define the probability that text $i$ is more biased than text $j$ to be

\begin{equation}
\label{eq:probmodel}
P(i \succ j) = \frac{e^{s_i}}{e^{s_i} + e^{s_j}},
\end{equation}
where $s_i, s_j \in \mathbb{R}$  are \textit{bias scores} of texts $i$ and $j$ respectively.
The higher the bias score of a text, the more biased it is.

To simplify the rest of the description, we assume that we are using static word embeddings.
It is straightforward to extend this to the case of contextual word embeddings, by using tokens in place of words.

We model the bias score of a text $i$ as the sum of the bias contributions of the words present in the text, weighted by the number of times each word occurs in the text.
More precisely, we define

\begin{equation}
\label{eq:docbias}
    s_i = \frac{1}{K_i} \sum_{w \in \mathcal{V}_i} n_i(w) B(w,i),
\end{equation}
where $B(w,i)$ is the bias contribution of the word $w$ given the topic of text $i$.
We also include a scaling factor $K_i$ to ensure that the bias score of a text does not depend on its length or generality. 
This enables us to compare the bias within a diverse set of texts.
We explicitly define $K_i$ later in this section.

We model the bias contribution $B(w,i)$ as a function of both the word $w$ and the text $i$, as the bias induced by words can change depending on the topic of the text.
For instance, the word \textit{malicious}, when used as an adjective to describe the nature of a specific person, usually indicates bias, but when used within a computer science article, it can be legitimate (e.g.,\textit{malicious code}).

To model this we define $B(w,i)$ as
\begin{equation}
\label{eq:bwi}
    B(w,i) = \mathbf{f_i}^T \mathbf{v_w},
\end{equation}
where $\mathbf{f_i} \in \mathbb{R}^d$ is the bias word \textit{query vector} for text $i$ and $\mathbf{v_w} \in \mathbb{R}^d$ is the embedding vector of word $w$.
The smaller the angle between $\mathbf{f_i}$ and $\mathbf{v_w}$ is, the higher the bias contribution of $w$ is, when it appears in text $i$.

The query vector $\mathbf{f_i}$ depends on the topic of text $i$.
We model it as
\begin{equation}
\label{eq:fi}
    \mathbf{f_i} = \mathbf{W}^T \mathbf{\widehat{t_i}}  + \mathbf{b},
\end{equation}
where $\mathbf{\widehat{t_i}}$ is the vector representation of text $i$, and $\mathbf{W} \in \mathbb{R}^{d \times d}$ and $\mathbf{b} \in \mathbb{R}^{d}$ are learned parameters.

Substituting (\ref{eq:fi}) in (\ref{eq:bwi}), and (\ref{eq:bwi}) in (\ref{eq:docbias}), and using (\ref{eq:ahat}) and (\ref{eq:t}) to simplify, we get the bias score of the text as

\begin{equation}
\label{eq:docbiasscore}
s_i = \frac{\lVert \mathbf{t_i} \rVert (\mathbf{\widehat{t_i}}^T \mathbf{W} \mathbf{\widehat{t_i}} + \mathbf{b}^T \mathbf{\widehat{t_i}})}{K_i}.
\end{equation}

We can see from (\ref{eq:t}) that the quantity $\lVert \mathbf{t_i} \rVert$ depends on the total number of words in the text.
If a text is concatenated with itself, $\lVert \mathbf{t_i} \rVert$ will increase even though the content and bias of the text do not change.

Also, if two texts $i$ and $j$ are similar (i.e., $\mathbf{\widehat{t_i}}$ and $\mathbf{\widehat{t_j}}$ have high similarity) and therefore should have similar bias, but $i$ is more specific and uses a less diverse set of words than $j$ (i.e., the embeddings $\mathbf{v_w}, \forall w \in \mathcal{V}_i$ have a higher mutual similarity than the embeddings $\mathbf{v_x}, \forall x \in \mathcal{V}_j$), then $\lVert \mathbf{t_i} \rVert$ tends to be larger than $\lVert \mathbf{t_j} \rVert$.
This could happen for instance if $j$ gives some context around the topic, placing it within a more general topic.

Since we would like the bias score of the text to not change in these cases, we define the scaling factor to be $K_i = \lVert \mathbf{t_i} \rVert$. We then have

\begin{equation}
\label{eq:docbiasscore2}
s_i = \mathbf{\widehat{t_i}}^T \mathbf{W} \mathbf{\widehat{t_i}} + \mathbf{b}^T \mathbf{\widehat{t_i}}.
\end{equation}

To interpret the model to identify the bias of words, we need to get the true values of all $B(w,i)$, for which we need precise inference to be possible for $\mathbf{W}$ and $\mathbf{b}$ (i.e., the model should be identifiable).
It is straightforward to see that this is satisfied if $\mathbf{W}$ is symmetric.
We therefore parameterize $\mathbf{W}$ as 

\begin{equation}
    \mathbf{W} = \mathbf{U} + \mathbf{U}^T,
\end{equation}

where $\mathbf{U} \in \mathbb{R}^{d \times d}$ is the variable that is optimized during learning.

While $B(w,i)$ gives the bias contribution of word $w$ when it appears in text $i$, we are also interested in obtaining the general bias score of a word in a given corpus of texts $\mathcal{C}$ without specifying any particular text.
Hence we define the general bias score of a word $w$ as an average of its bias score over all texts, i.e.,

\begin{equation}
    GB(w) = \frac{\sum_{i\in\mathcal{C}}B(w,i)}{\lvert \mathcal{C} \rvert} 
    = \mathbf{\bar{t}}^T \mathbf{W} \mathbf{v_w} + \mathbf{b}^T \mathbf{v_w},
\end{equation}

where 

\begin{equation}
    \mathbf{\bar{t}} = \frac{\sum_{i\in\mathcal{C}}\mathbf{t_i}}{\lvert \mathcal{C} \rvert}. 
\end{equation}

$GB(w)$ can be extended to the case of contextual word embeddings by averaging the bias score over each occurrence of a word.
However in this work we restrict it to models using static embeddings as it is significantly easier to compute in that case.




We call a version of our model including only the linear term $\mathbf{b}$ in (\ref{eq:fi})  as the \textit{Linear} model, and the full model including both terms as the \textit{Quadratic} model.

\subsection{Training}
We use the WAND dataset for training.
We split the revision pairs into train, validation and test sets in the ratio 90:5:5.
To avoid data leakage, we take care to ensure that all pairs from a given article are present in the same split.

We train each model by maximising the likelihood of the training data, under the probability model in (\ref{eq:probmodel}).
More precisely, we solve the optimisation problem given by

\begin{equation}
   \max_{\mathbf{\theta}} \prod_{(i,j)\in \mathcal{D}} P(i \succ j | \theta), 
\end{equation}

where $\theta = \{\mathbf{W}, \mathbf{b}\}$ is the set of parameters to be learned, $(i,j)\in\mathcal{D}$ are the revision pairs in the train set ($i$ is the version before the edit, $j$ is the version after the edit), and $P(i \succ j | \theta)$ is the probability that $i$ is more biased than $j$ given the parameters $\theta$, modelled as in equation \ref{eq:probmodel}. 

We use mini-batch stochastic gradient ascent for the maximisation.
Models based on static embeddings take approximately 2 hours to train, while those using contextual embeddings take approximately 2 days to train on an NVIDIA Quadro RTX 6000 GPU. 
For the models using the pre-trained contextual embeddings, we keep the weights of the embedding model fixed during training (i.e. no fine-tuning) due to constraints on GPU usage time. 
We do not observe any overfitting based on the performance on the validation set and therefore do not use any regularisation.

\section{Evaluation and Applications}
\label{sec:eval}

In this section, we evaluate the performance of our models, examine their interpretability and explore their applications in a variety of domains.

\subsection{Evaluation}

We evaluate the generalization ability of our models on the task of pairwise bias classification by measuring their accuracy on the test set.
Since we are comparing the bias within a pair and a difference is always present within a pair (both versions do not have exactly the same bias), the concept of type I and type II errors are not as relevant and accuracy is a sufficiently informative metric.

We compare against several baselines which we describe below:
\begin{itemize}
    \item \textbf{Random}: The random classifier predicts one of the two versions in a pair uniformly at random to be the more biased one.
    \item \textbf{Wiki Words to Watch (Words2Watch)}: Wikipedia maintains a list of words that could potentially cause bias called \textit{Words to Watch} \cite{wiki2022w2w}. 
    This classifier first compares the count of such words among the two versions in the pair.
    If the counts are equal, it picks one of two versions uniformly at random.
    Otherwise, it predicts the version having the higher count to be the more biased version.
    \item \textbf{Sentence Bias Aggregate (SentAgg)} These classifiers are based on the sentence-level bias models developed in \cite{zhong2021wikibias}.
    
    We first extract a dataset of biased and neutral sentences from the revision pairs in our train set, following the procedure in \cite{pryzant2020wnc}. 
    Like in \cite{zhong2021wikibias}, we then train sentence-level classifiers based on the pre-trained BERT model \cite{devlin2019bert} on this dataset to predict if a sentence is biased or neutral.
    
    To perform the version level bias comparison for a pair, SentAgg aggregates the predicted probability for the sentences in each version and compares the aggregated probability of the two versions.
    
    We construct three versions of the SentAgg classifier: 
    \begin{itemize}
    \item \textit{SentAgg-Max}, where BERT weights are not finetuned when the sentence classifier is trained and aggregated probability for a version is computed as the \textit{maximum} of the predicted probabilities for the sentences in the version, 
    \item \textit{SentAgg-Mean}, where BERT weights are not finetuned but the aggregated probability is computed as the \textit{mean} of the predicted probabilities and
    \item \textit{SentAgg-FT-Mean}, where BERT weights are \textit{finetuned} when the sentence classifier is trained and aggregated probability is computed as the \textit{mean} of the predicted probabilities.
    \end{itemize}

\end{itemize}

The test accuracy of all baseline models and our models, and their 95\% confidence intervals are given in Table \ref{tab:results}.
The accuracies of the top two models are highlighted in bold.
We also include a human performance benchmark which was obtained by one of the authors manually labelling 100 randomly chosen pairs from the test set.

\begin{table}[]
\centering
\begin{tabular}{@{}ll@{}}
 \textbf{Model}   & \textbf{Accuracy(\%)}  \\ 
 \midrule
Random            &    $50\pm0.46$             \\
Words2Watch    &    $63.4\pm0.44$                 \\
SentAgg-Max       &    $55.33\pm0.46$            \\
SentAgg-Mean      &    $68.35\pm0.43$            \\
SentAgg-FT-Mean     &   $76.01\pm0.39$                   \\
\midrule
Static Linear     &    $75.29\pm0.40$ \\
Static Quadratic  &    $\mathbf{76.84\pm0.39}$ \\             Contextual Linear &    $74.35\pm0.40$  \\
Contextual Quadratic &  $\mathbf{77.56\pm0.38}$ \\
\midrule
Human             &     $74.00\pm8.60$
\end{tabular}
\caption{Accuracy of models}
\label{tab:results}
\end{table}

The best performance of $77.56\%$ accuracy is achieved by our \textit{Quadratic} model using contextual word embeddings, and the same model using static word embeddings achieves a close second with an accuracy of $76.84\%$.
Both models significantly exceed the performance of all baselines.

The higher accuracy achieved by our $Quadratic$ models relative to our $Linear$ models suggests that the information given by the document topic in computing $B(w, i)$ is beneficial.

The best-performing baseline is \textit{SentAgg-FT-Mean} which uses a fine-tuned BERT model.
Note that our models achieve a better performance than it does despite not using fine-tuning.

The large difference in performance between \textit{SentAgg-Mean} and \textit{SentAgg-FT-Mean} suggests that fine-tuning could further improve the performance of our models to some extent, albeit with much higher computational costs.

We now compare the models in terms of their inference time, i.e., the time taken for the trained model to score and compare a pair of versions.

\begin{table}[]
\centering
\begin{tabular}{@{}ll@{}}
 \textbf{Model}   & \textbf{Inference time(ms)}  \\ 
 \midrule
Static Quadratic (on CPU)  &    $130$ \\         
Contextual Quadratic &  $816$ \\
SentAgg-FT-Mean     &  $1,278$      \\    
\end{tabular}
\caption{Inference time of the best-performing models per version pair, averaged over 1,000 randomly chosen pairs from the test set.}
\label{tab:inftime}
\end{table}

Remarkably, our \textit{Static Quadratic} model outperforms \textit{SentAgg-FT-Mean} and achieves a performance similar to \textit{Contextual Quadratic} while requiring much less computational resources than both those models for training and inference.
From Table \ref{tab:inftime}, we see that inference in \textit{Static Quadratic} is almost an order of magnitude faster than \textit{SentAgg-FT-Mean}, while also not using a GPU.

In addition to being fast and accurate, the \textit{Static Quadratic} model is also highly interpretable as it doesn't use deep neural networks.
In the experiments that follow where we illustrate our model's interpretability and its application in diverse domains, we use the \textit{Static Quadratic} model unless mentioned otherwise.

\subsection{Interpretation}
A salient feature of the model is its ability to provide explanations for its bias scoring, by computing scores for individual words in the text.
We interpret the trained model to see the words indicative of bias.
First, we obtain the general bias score $GB(w)$ for every word $w$ in the WAND dataset.
The list of top $30$ words with the highest $GB(w)$, and the list of $10$ words at the $10^{th}$ percentile are given in Table \ref{tab:generalbias}.

\begin{table}[]
\centering
\begin{tabular}{cccc}
\multicolumn{3}{c}{\textbf{High} $GB(w)$}                & \textbf{Low} $GB(w)$         \\ \hline
\textbf{1-10} & \textbf{11-20} & \textbf{21-30} & $\mathbf{P_{10\%}}$ \\ \hline
impressive        & stunning       & spectacular         & waived               \\
finest        & horrible         & arrogant       & readings         \\
superb      & splendid      & memorable  & discussed             \\
wonderful      & talented          & awesome    & convened        \\
toughest      & amazing       & magnificent       & attended               \\
formidable       & pleasing     & ruthless        & supplements               \\
brilliant        & proud        & daring       & chaired           \\
exciting    & fascinating      & greatest       & grams            \\
beautiful       & clever       & courageous     & served             \\
excellent     & terrible       & incredible  & suggested             
\end{tabular}
\caption{Words $w$ in decreasing order of $GB(w)$}
\label{tab:generalbias}
\end{table}

We see that the words with the highest scores are typically subjective adjectives and other subjective words.
The words with lower scores are typically verbs and common nouns.

\begin{figure}[h]
  \centering
  \includegraphics[width=0.9\columnwidth]{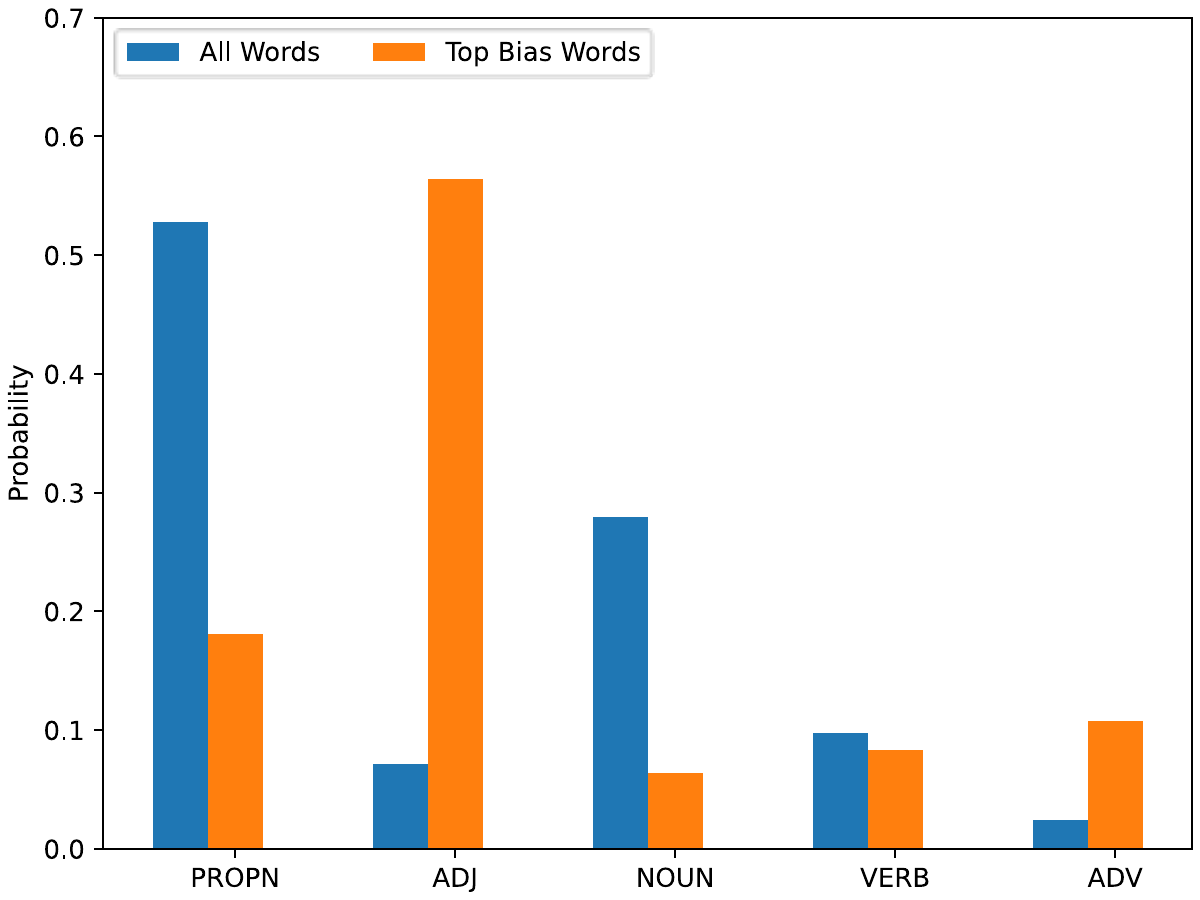}
  \caption{Comparison of POS distributions}
  \label{fig:poscomp}
\end{figure}

To have a more comprehensive analysis, we plot in Figure \ref{fig:poscomp} the part-of-speech (POS) distribution of the top 1,000 words in terms of $GB(w)$ in comparison to that of all words.
We see clearly that the proportions of adjectives (ADJ) and adverbs (ADV) in the bias-inducing words are significantly higher than that of all words, while the proportion of proper nouns (PROPN) and common nouns (NOUN) are significantly lower.
The proportion of verbs (VERB) is nearly the same. 

To provide external validation for the word bias scores $GB(w)$ generated by the model we rely on the Wikipedia \textit{Words to Watch} list.
In Table \ref{tab:biasvalid}, we give the mean $GB(w)$ of all words as well as the words in the \textit{Words to Watch} list, with their 95\% confidence intervals.
We see clearly that mean $GB(w)$ of \textit{Words to Watch} is significantly higher than that of all words.

\begin{table}[]
\centering
\begin{tabular}{@{}ll@{}}
\textbf{Word Type}      & \textbf{Mean $GB(w)$} \\ \midrule
All & $48.84  \pm  0.28$ \\ 
Words2Watch & $108.21  \pm  14.30$  \\
\end{tabular}
\caption{Mean $GB(w)$ of all words vs Words2Watch}
\label{tab:biasvalid}
\end{table}

We can also compare the values of $B(w,i)$ for the same word in different articles to see how the bias induced by the word changes depending on the article's topic.
For instance, the word \textit{poorly} when used in the sense of bad performance in sports (in the article \textit{Howard Johnson} (baseball player)) has a $B(w,i)$ score of 324.12. In contrast, it has a much lower bias score of 30.56 when used to describe something `burning poorly' in the article \textit{Hydrogen Storage}. 

The \textit{Contextual Quadratic} model can also be interpreted to identify words and especially multi-word phrases that induce bias.
An example is shown in Table \ref{tab:incontext}, where the model correctly identifies the bias-inducing phrase \textit{without a doubt}, which is also mentioned as part of Wikipedia's \textit{Words to Watch}.
The \textit{Static Quadratic} model fails to identify the phrase and incorrectly identifies \textit{amp} to be a bias word.

\begin{table}[]
\centering
\begin{tabular}{|p{8cm}|}
\hline
Another change was that apart from no drummer appearing on the album all guitars were recorded directly into the mixing desk without a guitar \underline{amp}. This is \textbf{without a doubt the most} \underline{brutal} album \textbf{ever} made without a drumkit and guitar \underline{amp}. The spontaneity brought the focus away from feats of musicianship and sent it towards \underline{monstrous sounding riffs} and \underline{\textbf{great}} songs. \\
\\
\hline
\end{tabular}
\caption{An excerpt from the article \textit{The Berzerker}, a death metal band. Words with the highest bias according to the \textit{Contextual Quadratic} model are highlighted in bold. The highest bias words according to the \textit{Static Quadratic} model are underlined.}
\label{tab:incontext}
\end{table}

We now comment on some applications of our model for scoring bias in different settings.
Note that the model has only been trained on Wikipedia data.
In each case, we apply the same preprocessing steps to the text as we did while training.

\subsection{Bias in Wikipedia}
\label{sec:wikibias}
We first apply the model to its training domain, namely scoring the bias in Wikipedia articles.
We use the data in Wikipedia: Controversial Issues dataset for the analysis in this section.

\subsubsection{Article-level bias}
First, we compute the average bias score of each article across its revisions and identify the articles with the highest and lowest scores.
The results for the articles within the \textit{Politics and Economics} section of the dataset are given in Table \ref{tab:MostLeastArticles}.
We see that the articles with the highest scores are about subjective topics like different '-ism's, and highly controversial topics like racism and denial of genocide.
By comparison, the articles with the lowest scores tend to be about fairly objective topics (although still controversial, as we are comparing within the list of controversial topics) like Macedonia, CBC News, and the National Rifle Association. 
The article on Russian interference in US elections, although it deals with a controversial topic, is well-sourced and protected.

\begin{table}[]
\centering
\begin{tabular}{@{}cc@{}}
\textbf{Highest mean $s_i$}  & \textbf{Lowest mean $s_i$}   \\ \midrule
Anti-Italianism                                & Macedonia                                   \\ 
Patriotism                           & National Rifle Association                                    \\
Anti-Irish racism                           & CBC News                           \\
Genocide denial                          & Federal Marriage Amendment                     \\
Black Supremacy                        & Russian Interference...                                  
\end{tabular}
\caption{Most and least biased articles in the \textit{Politics and Economics} section}
\label{tab:MostLeastArticles}
\end{table}

\subsubsection{Topic-level bias}
\begin{figure}
  \centering
  \includegraphics[width=0.9\columnwidth]{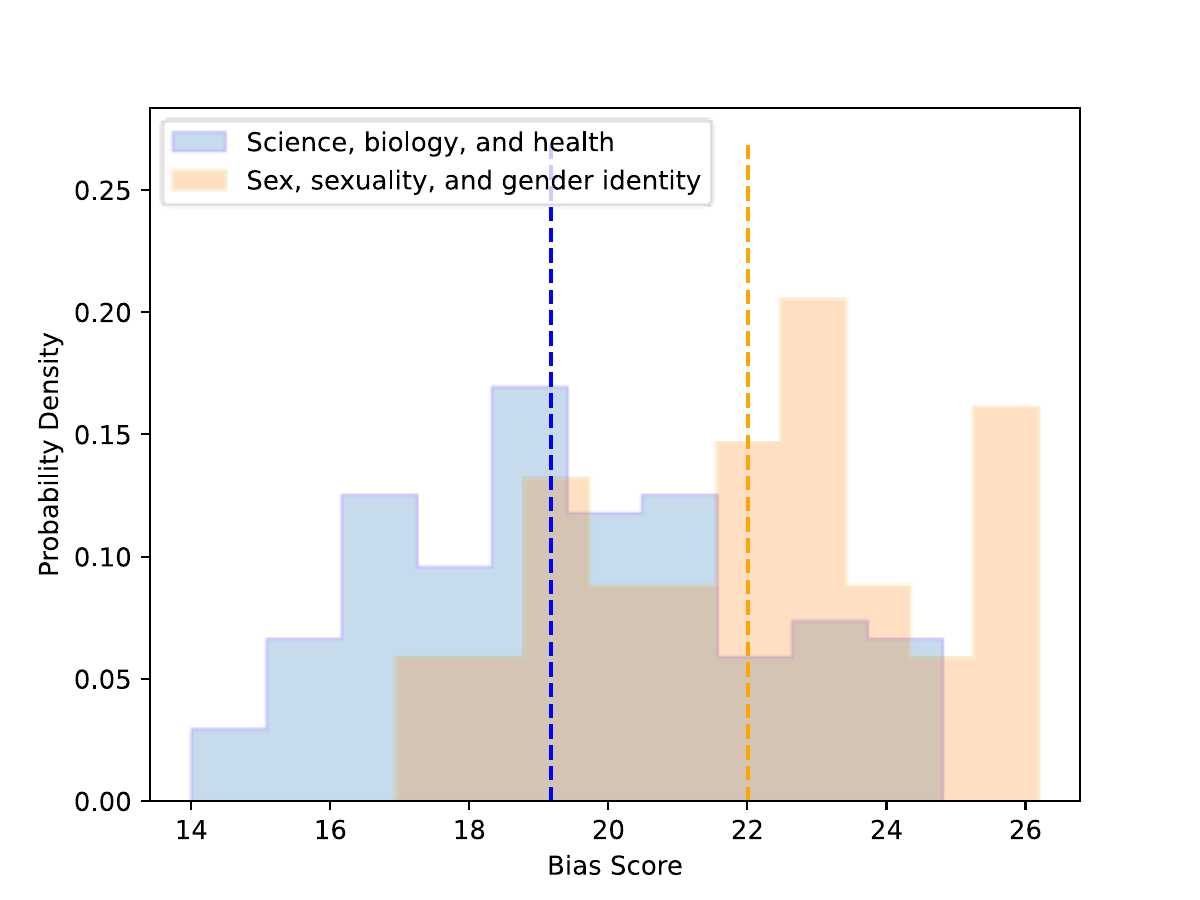}
  \caption{Distribution of bias scores within topics.}
  \label{fig:wikicontro}
\end{figure}

Second, we compare the distributions of bias scores of articles in two different topics, namely \textit{Science, biology and health}, a relatively objective topic, and \textit{Sex, sexuality and gender identity} which contains articles on highly controversial topics such as gay rights. The distributions are given in Figure \ref{fig:wikicontro}. The vertical bars show the positions of the means.

We see that the articles in the \textit{Sex} topic generally have a higher bias score as expected.
There is some overlap as many articles such as \textit{Abortion}, \textit{AIDS}, etc. occur in both topics.

\subsubsection{Temporal Evolution of Bias}
\label{sec:tempevolution}
\begin{figure}[h]
  \centering
  \includegraphics[width=0.9\columnwidth]{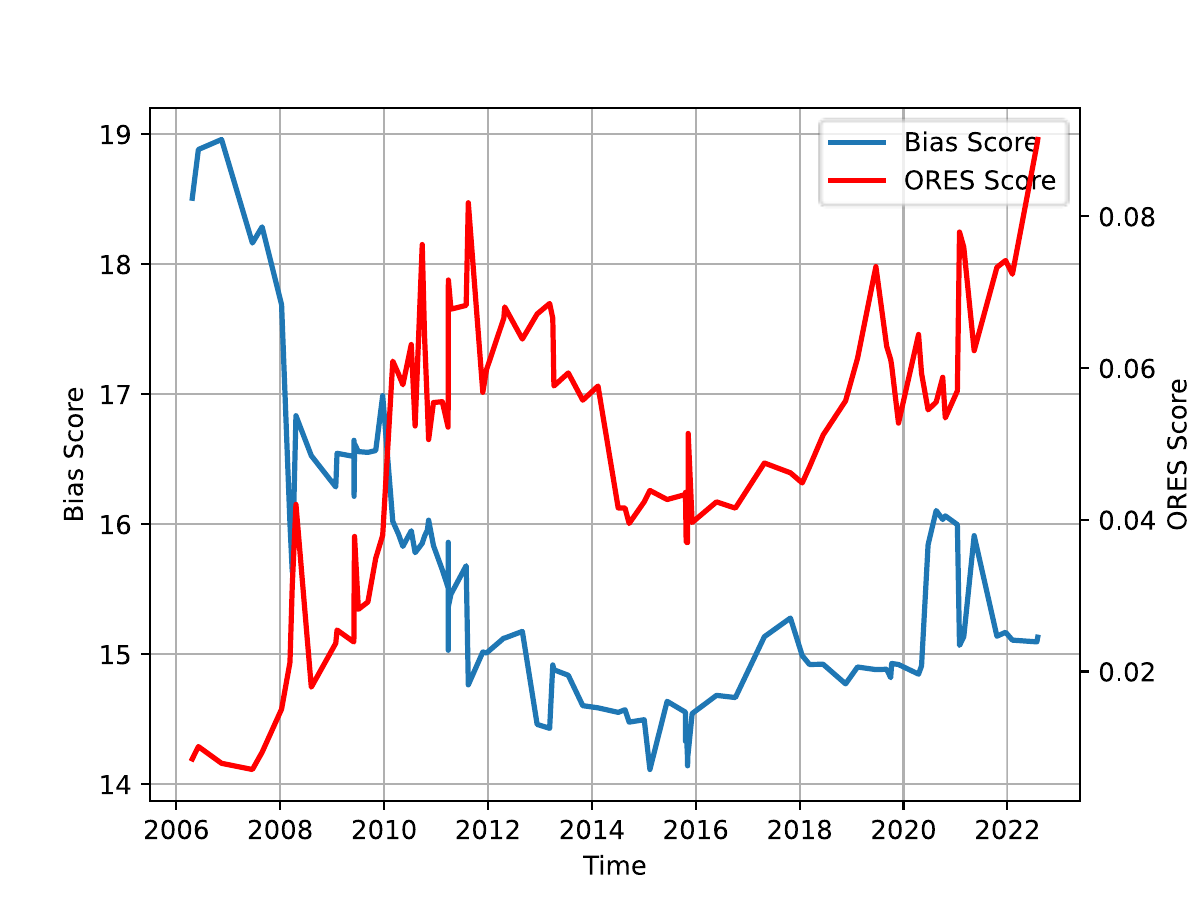}
  \caption{Bias score of \textit{Heritability of IQ} over time. Wikipedia ORES article quality scores are plotted for comparison. Spearman correlation: -0.27, p-value 0.008}
  \label{fig:hiq}
\end{figure}

Third, we study the evolution of bias over time by plotting the bias score $s_i$ over time as revisions are made to an article.
As examples we show the plot of the article \textit{Heritability of IQ} in Figure \ref{fig:hiq}. 

For comparison, we also plot in the same figure the article quality score computed by the Wikipedia Objective Revision Evaluation Service (ORES) \cite{halfaker2020ores}. 
ORES uses a machine learning model to predict article and edit quality based primarily on structural features (templates, headings, links, citations etc.) and some textual features (length, Words to Watch matches etc.).
Wikipedia editors use the ORES article quality scores to periodically evaluate articles and identify the ones to focus on for editing (such as popular articles that have low quality).

We refer to the probability that ORES gives for an article to be a `Good Article' as per Wikipedia's criteria \cite{wikipedia2023GA} as the \textit{ORES score}.
Importantly, one of the criteria is that the article should be neutral as per \cite{wikipedia2022npov}.
The ORES score for a revision can be obtained by querying a public API \cite{wikipedia2023ores}.

We see from Figure \ref{fig:hiq} that the bias score computed by our model has a negative correlation with the ORES score, which is expected as bias negatively affects quality.
The median Spearman correlation across all articles in the dataset is -0.27 and the example in Figure \ref{fig:hiq} has the same correlation.

Note that we don't expect a very high magnitude of correlation since ORES also considers other aspects of article quality besides bias and primarily uses structural features which our model doesn't use. 
The advantage of our model is that it does a much better assessment of the bias of the textual part, as it is not restricted to a manually built lexicon of bias words unlike ORES (cf. the performance of Words2Watch baseline in Table \ref{tab:results}).
Also, the fact that our model only uses textual features enables it to generalise to other domains as we show in subsequent sections.

We manually examined the text of the revisions of \textit{Heritability of IQ} at different points in time to examine the reason behind the bias trend seen in Figure \ref{fig:hiq}. 
We see that the bias is relatively high in the beginning as the article has been written primarily by a single author and lacks a balanced view.
Over time other authors incorporate competing views and rephrase statements to reflect the presence of controversy.
The wording is also improved to make it sound more objective without changing the meaning (eg: \textit{stated} instead of \textit{claimed}).
This causes the bias score to generally decrease over time.
However, there are instances where an editor inserts or deletes a large amount of biased text that causes sharp fluctuations in the bias score as seen in Figure \ref{fig:hiq}.
In most cases though, this is quickly reversed by other editors.
We believe this behaviour is generally followed in the case of other articles as well, based on a quick perusal of their revision histories.



In addition to the evolution of bias for individual articles, we also study the trend of the average bias across articles over time.
This would help to answer questions such as whether on average the bias of an article decreases over time in Wikipedia (RQ3), and if so how fast does it decrease.

\begin{figure}[h]
  \centering
  \includegraphics[width=0.9\columnwidth]{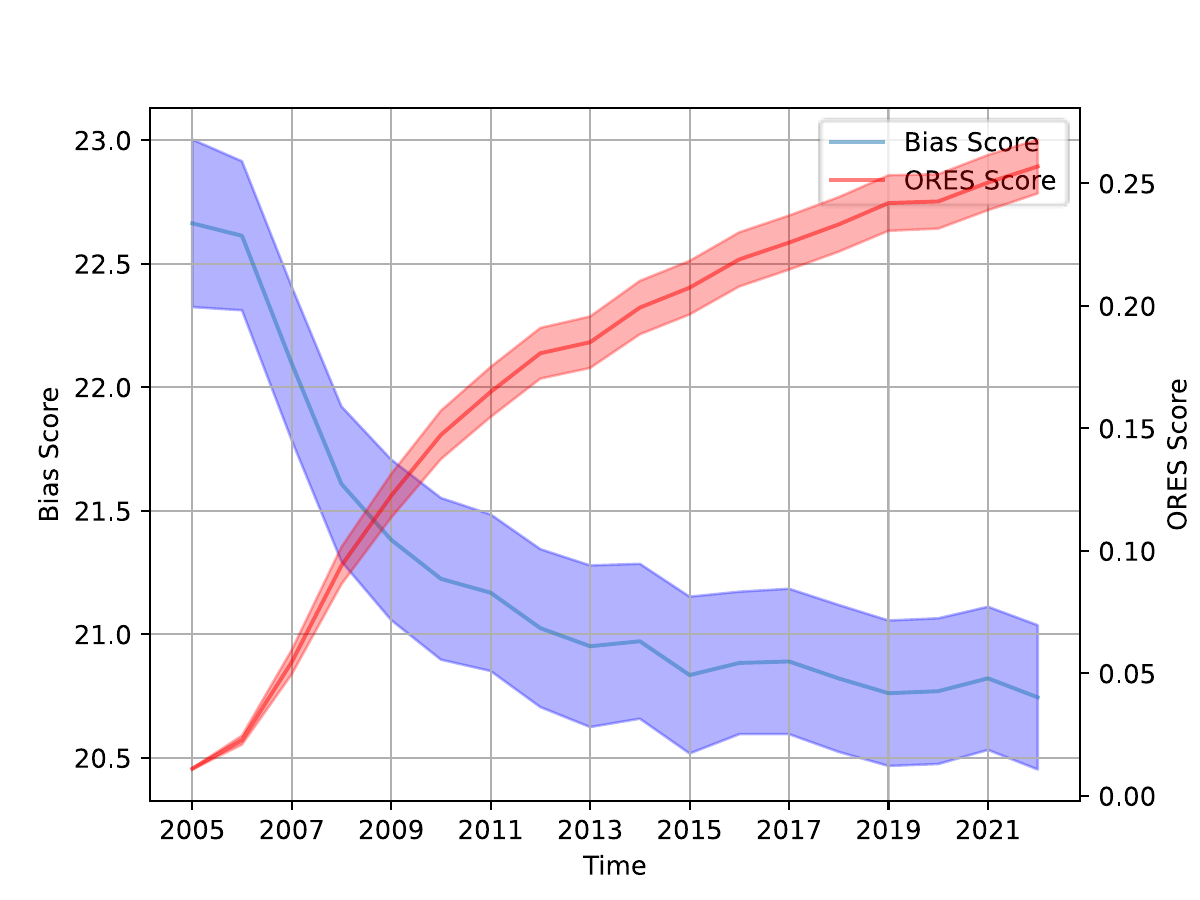}
  \caption{Average bias score and ORES score of articles over time.}
  \label{fig:biasdecrease}
\end{figure}

We consider all articles in the dataset that were created around the same time (in 2003 or 2004), and average each of their bias scores at the same points in time throughout their history.
We get the trend shown in Figure \ref{fig:biasdecrease}, where the dark line is the average bias score and the shaded area indicates the 95\% confidence interval.

We can clearly see that on average the bias of an article decreases over time until it reaches a steady state, and that it reaches this state in about ten years.
The increasing trend of the ORES score also supports this conclusion.

\subsection{Media Bias}
We now apply the model to score bias in the domain of news media, a different domain from its training domain of Wikipedia.
We use the News Dataset in this analysis and we treat the text content of each news article as a document.

First, we try to estimate the relative bias level of different outlets to see if we can rank them and identify the ones that are most and least biased.
We note that some of the outlets in the dataset, despite having clearly recognised political leanings, also publish many articles verbatim from news agencies such as Reuters and Associated Press.
Including such articles would unfairly lower the bias score for these outlets as the articles from news agencies are known to use objective language.
We, therefore, remove such articles from the analysis (except for the articles published by the news agencies themselves).  
We then obtain a bias score for the remaining articles using our trained model.

We are primarily interested in the relative bias of left and right outlets as compared to the center.
Hence we include only the stories that have at least one article from a center outlet.
Then for every news story, we order the articles covering the story in terms of the bias score, and compute the percentile bias score for each article in the story.
Finally, we compute the average of the percentile bias score of the articles from a news outlet to get the mean percentile bias score of the outlet. 

\begin{figure}
  \centering
  \includegraphics[width=0.9\columnwidth]{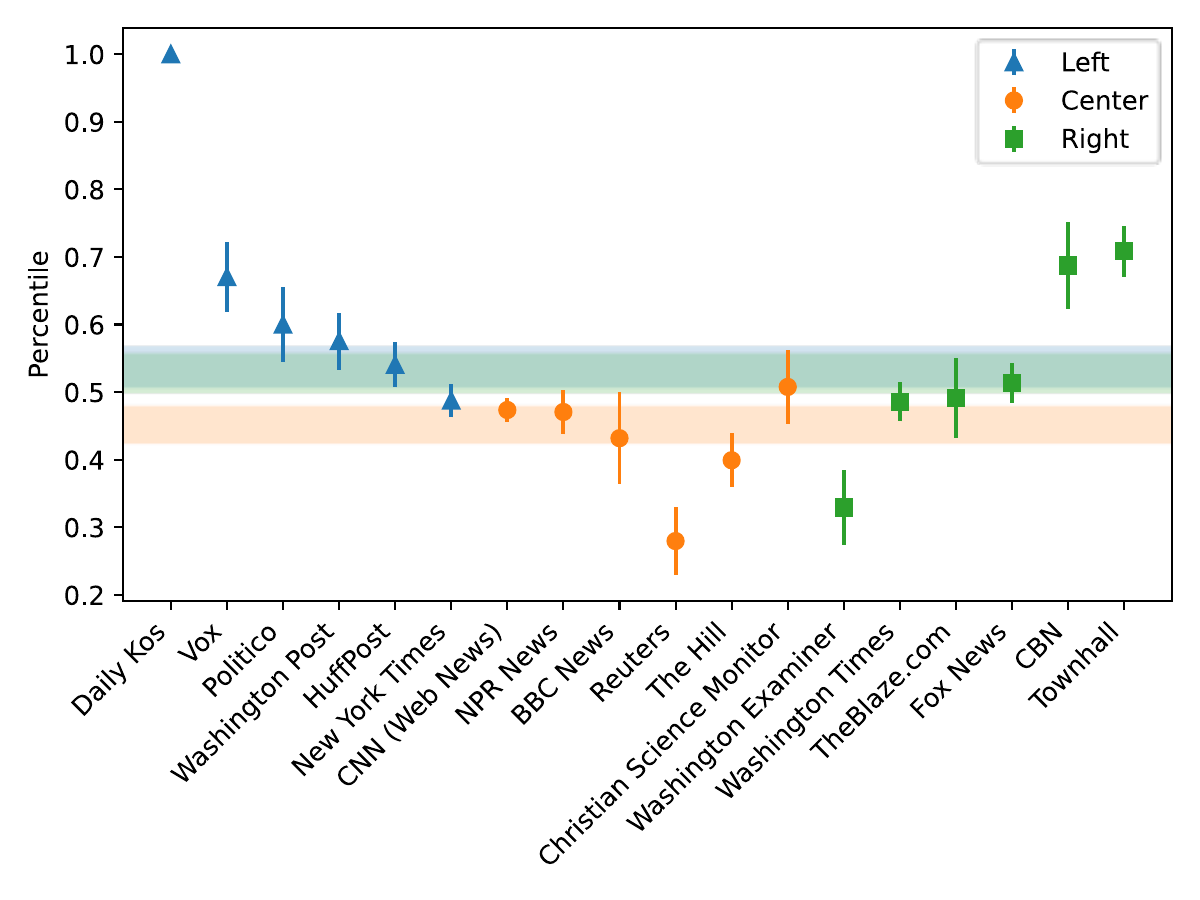}
  \caption{Mean percentile bias score of news outlets. The bands show the confidence interval of the mean percentile bias score for left, right and center articles. These include articles from outlets not shown in the figure.}
  \label{fig:Ushape}
\end{figure}

We plot the mean percentile bias scores of the outlets along with their 95\% confidence intervals in Figure \ref{fig:Ushape}.
For clarity, we only show in the plot the 6 outlets with the smallest confidence interval from each category (left, right and center).
We also show the confidence intervals of the mean percentile bias scores of left, right and center articles, including those from outlets not shown in the figure.

Although there is overlap between individual outlet scores, we see from the confidence intervals of the mean scores that articles from center outlets have significantly lower mean score than those from left and right outlets.
Looking at the individual outlet scores, we see that the \textit{Reuters} news agency which is known for its policy of objective language has the lowest bias score.
\textit{The Hill} which claims to provide ``objective'' and ``non-partisan coverage" also has a relatively low bias score.
On the other hand, outlets like \textit{Daily Kos} on the liberal side and \textit{Townhall} on the conservative side are open about their political bias.
Their articles commonly include partisan commentary on news events and consequently have a very high bias score.
This is also true for outlets such as \textit{Vox} and \textit{Christian Broadcasting Network (CBN)} that are relatively less open about their bias.
Mainstream outlets such as \textit{New York Times} and \textit{CNN} have a similar and moderate level of bias.
\textit{Washington Examiner} is an outlier; it is considered by AllSides to have a Lean-Right bias but has a quite low mean bias score.
On manually examining their articles in our dataset, we find that bias appears here in the form of giving a greater fraction of coverage to certain views, rather than word choice or other forms of subjective language.
Our model is not expected to detect such forms of bias which explains the low bias score.

Finally, we plot the distribution of bias scores of all the news articles in Figure \ref{fig:threedist}, along with the distribution of scores in Wikipedia. 
We see that the scores are generally higher, as news articles frequently contain subjective commentary on events as discussed earlier, while this is disallowed in Wikipedia.  

\subsection{Bias in Legal Texts}

We use the European Parliament Amendments dataset to study bias scoring in the legal domain.
We give in Table \ref{tab:legalbias} the mean bias scores of the different subsets of legal texts in the dataset.

\begin{figure}
\centering
\includegraphics[width=0.9\columnwidth]{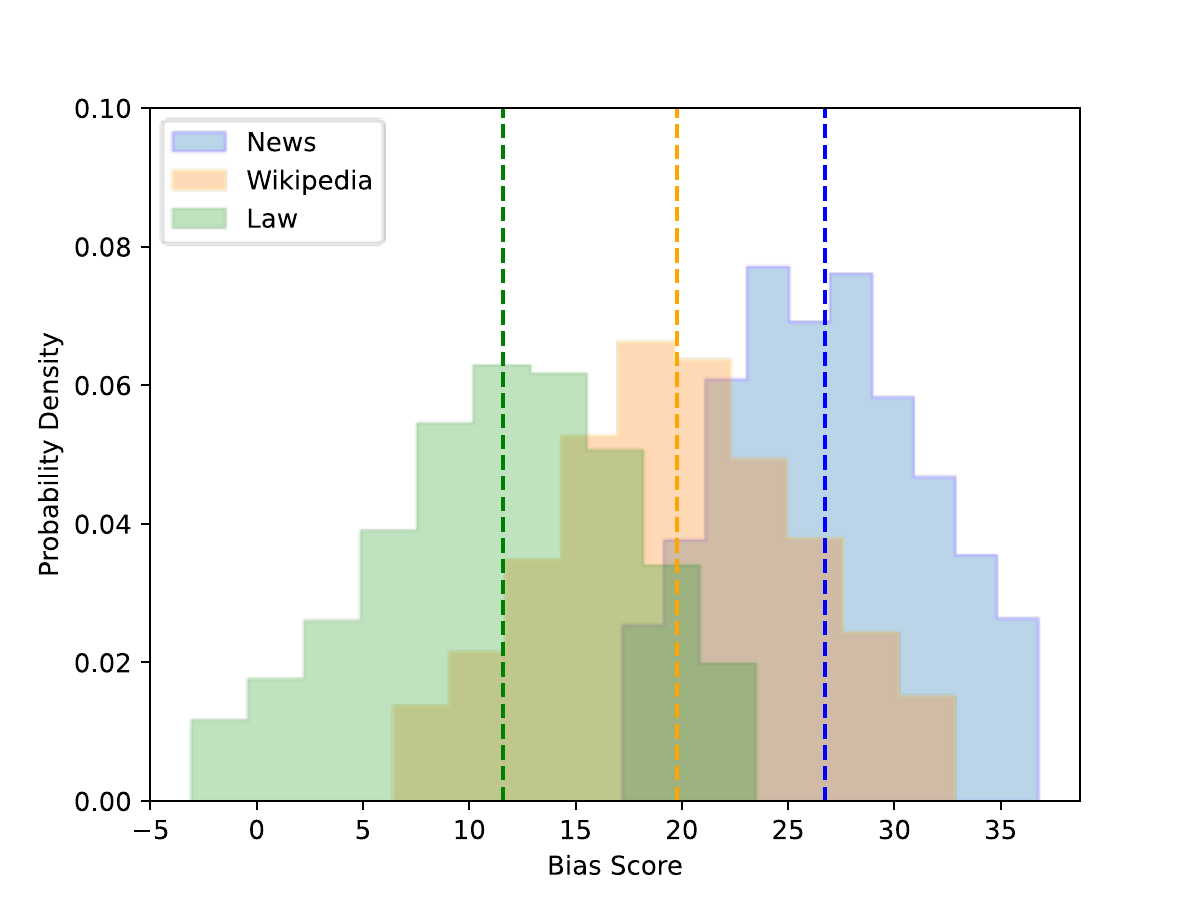}
  \caption{Distribution of bias scores across domains}
  \label{fig:threedist}
\end{figure}

First, we see that the magnitude of bias scores is significantly lower than that of Wikipedia, as is also clear from the distribution of bias scores in Figure \ref{fig:threedist}.
This is the opposite of what was observed in the case of News.
This could be due to the fact that legal texts are carefully crafted to be objective so as to minimise ambiguity in interpretation of the law.
They also tend to avoid partisan language, especially in the introduction sections, so that the text can be accepted by a broad set of legislators who may have diverse viewpoints. 

Interestingly, we see that the average bias of the amendments that the parliamentarians propose is higher than that of the original text proposed by the commission.
On manually examining the amendments with the highest difference in bias scores, we see that many of them change the introductory sections of the law (explanatory memoranda, recitals etc.) by introducing partisan and subjective language. 
An example is given in Table \ref{tab:lawincontext}.
Nevertheless, we see that among the proposed amendments, the ones that get accepted have relatively a smaller bias on average.

\begin{table}[]
\centering
\begin{tabular}{|p{8cm}|}
\hline
member states \st{may benefit from} \textit{have not benefitted from} support in addressing challenges as regards the design and implementation of structural reforms . \st{these challenges may be dependent on various factors , including limited administrative and \textbf{institutional} capacity or inadequate application and implementation of union legislation .}
\textit{on the contrary, the reforms have massively swelled the ranks of the unemployed and helped social inequalities to take hold within the eu , while leaving increasing numbers of citizens \textbf{destitute} by the day .}
\\
\\
\hline
\end{tabular}
\caption{A proposed amendment to the law on \textit{Establishment of the Structural Reform Support Programme for the period 2017 to 2020}. The most biased word in the original text and added text (italicized), is highlighted in bold. The bias score increased from -0.41 to 19.61}
\label{tab:lawincontext}
\end{table}

\begin{table}[]
\centering
\begin{tabular}{@{}ll@{}}
\textbf{Legal text}      & \textbf{Mean Score} \\ \midrule
All (Original + Amendments)      & $11.34  \pm  0.04$ \\ 
Original            & $9.70  \pm  0.10$ \\
Amendments         & $11.81  \pm  0.05$ \\
Amendments with at least 1 edit accepted & $11.64  \pm  0.08$ \\
Amendments with at least 1 edit rejected              & $12.01  \pm  0.06$ \
\end{tabular}
\caption{Mean bias score of legal texts.}
\label{tab:legalbias}
\end{table}

\section{Conclusion}
\label{sec:conclusion}
In this paper, we developed a simple, interpretable model of bias in documents.
The model combines elements of discrete-choice theory and metric word embeddings.
We curated two novel datasets based on Wikipedia revision histories to train and evaluate our model.
The model is trained on pairs of revisions of Wikipedia articles, where one revision was corrected for POV issues by another edit.
Formulating the problem as identifying the larger bias in a pair, rather than estimating absolute bias, reduces subjectivity and issues of thresholding.
We obtain strong performance on a holdout set of pairs of Wikipedia revisions.
Importantly, the model is interpretable: we can score individual words, a feature that an editor might rely upon to quickly identify the most problematic parts of a document that contribute to the bias.
The list of globally most biased words contains a convincing list of strong adjectives (often in their superlative form), as well as terms that tend to express strong emotions.
We explored the predictions of the model over a dataset of news articles and over law amendments.
The bias distributions over the three domains (Wikipedia, news, laws) are quite different, with news the most biased, and laws the least, which can be explained by the policies governing the creation of content in each of them.
We also observe that we can score the bias in different news outlets; these scores align well with crowdsourced labelings of bias of these outlets.
Ultimately, we hope that this work will contribute to better identify and correct both deliberate and subconscious bias in online discourse.

\textbf{Broader Impact and Ethical Considerations}. In addition to the bias scoring model we developed, which is applicable in a wide variety of domains, the methodology that we adopted of casting bias as a relative quantity and learning from pairwise comparisons can be extended to a much broader set of problems in natural language processing.
It is particularly suited to those settings where the threshold for absolute categorisation may be subjective or depends on many factors, while there is more agreement in comparisons. 
Examples include measuring hateful content, agreeableness, humor, sentiment etc. 

All data we use in this work is from publicly available sources.
Wikipedia data that we collect is publicly released under the CC BY-SA and GFDL licenses and analysis of this content does not require informed consent.

Machine learning models are limited by the data that they learn from.
Therefore our models inherit any bias that is inherent in Wikipedia's neutrality policy or the manner in which the editors interpret and enforce that policy.
An editorial decision that is made based on the output of these models could also serve to reinforce such bias.
However, the interpretability of our models mitigate this risk to some extent.
For instance, if the model generates an unexpected output an editor can obtain the words that contributed to the model's assignment of a high or low bias score and perform an informed reassessment. 

\bibliography{aaai22}

\begin{thebibliography}{29}
\providecommand{\natexlab}[1]{#1}

\bibitem[{{Allen Institute for AI}(2022)}]{allenai2022longformer}
{Allen Institute for AI}. 2022.
\newblock longformer-base-4096.
\newblock \url{https://huggingface.co/allenai/longformer-base-4096}.
\newblock Accessed: 2022-08-1.

\bibitem[{AllSides(2022)}]{allsides2022allsides}
AllSides. 2022.
\newblock AllSides.com.
\newblock \url{https://www.allsides.com/unbiased-balanced-news}.
\newblock Accessed: 2022-09-15.

\bibitem[{Aroyo et~al.(2019)Aroyo, Dixon, Thain, Redfield, and
  Rosen}]{aroyo2019crowd}
Aroyo, L.; Dixon, L.; Thain, N.; Redfield, O.; and Rosen, R. 2019.
\newblock Crowdsourcing Subjective Tasks: The Case Study of Understanding
  Toxicity in Online Discussions.
\newblock In \emph{Companion Proceedings of The 2019 World Wide Web
  Conference}, WWW '19, 1100–1105. New York, NY, USA: Association for
  Computing Machinery.
\newblock ISBN 9781450366755.

\bibitem[{Beltagy, Peters, and Cohan(2020)}]{Beltagy2020Longformer}
Beltagy, I.; Peters, M.~E.; and Cohan, A. 2020.
\newblock Longformer: The Long-Document Transformer.
\newblock \emph{arXiv:2004.05150}.

\bibitem[{Bojanowski et~al.(2017)Bojanowski, Grave, Joulin, and
  Mikolov}]{bojanowski2017enriching}
Bojanowski, P.; Grave, E.; Joulin, A.; and Mikolov, T. 2017.
\newblock Enriching Word Vectors with Subword Information.
\newblock \emph{Transactions of the Association for Computational Linguistics},
  5: 135--146.

\bibitem[{Bradley and Terry(1952)}]{bradley1952comparison}
Bradley, R.~A.; and Terry, M.~E. 1952.
\newblock Rank Analysis of Incomplete Block Designs: I. The Method of Paired
  Comparisons.
\newblock \emph{Biometrika}, 39(3/4): 324--345.

\bibitem[{Chen et~al.(2018)Chen, Wachsmuth, Al-Khatib, and
  Stein}]{chen2018bias}
Chen, W.-F.; Wachsmuth, H.; Al-Khatib, K.; and Stein, B. 2018.
\newblock {Learning to Flip the Bias of News Headlines}.
\newblock In Gatt, A.; Goudbeek, M.; and Krahmer, E., eds., \emph{11th
  International Natural Language Generation Conference (INLG 2018)}, 79--88.
  Tilburg, Netherlands: Association for Computational Linguistics.

\bibitem[{De~Kock and Vlachos(2022)}]{de-kock-vlachos-2022-leveraging}
De~Kock, C.; and Vlachos, A. 2022.
\newblock Leveraging {W}ikipedia article evolution for promotional tone
  detection.
\newblock In \emph{Proceedings of the 60th Annual Meeting of the Association
  for Computational Linguistics (Volume 1: Long Papers)}, 5601--5613. Dublin,
  Ireland: Association for Computational Linguistics.

\bibitem[{Devlin et~al.(2019)Devlin, Chang, Lee, and
  Toutanova}]{devlin2019bert}
Devlin, J.; Chang, M.-W.; Lee, K.; and Toutanova, K. 2019.
\newblock {BERT}: Pre-training of Deep Bidirectional Transformers for Language
  Understanding.
\newblock In \emph{Proceedings of the 2019 Conference of the North {A}merican
  Chapter of the Association for Computational Linguistics: Human Language
  Technologies, Volume 1 (Long and Short Papers)}, 4171--4186. Minneapolis,
  Minnesota: Association for Computational Linguistics.

\bibitem[{Halfaker and Geiger(2020)}]{halfaker2020ores}
Halfaker, A.; and Geiger, R.~S. 2020.
\newblock Ores: Lowering barriers with participatory machine learning in
  wikipedia.
\newblock \emph{Proceedings of the ACM on Human-Computer Interaction},
  4(CSCW2): 1--37.

\bibitem[{Keegan and Fiesler(2017)}]{keegan2017evolution}
Keegan, B.; and Fiesler, C. 2017.
\newblock The evolution and consequences of peer producing Wikipedia's rules.
\newblock In \emph{Proceedings of the International AAAI Conference on Web and
  Social Media}, volume~11, 112--121.

\bibitem[{Kristof et~al.(2021)Kristof, Suresh, Grossglauser, and
  Thiran}]{kristof2021war}
Kristof, V.; Suresh, A.; Grossglauser, M.; and Thiran, P. 2021.
\newblock War of Words II: Enriched Models of Law-Making Processes.
\newblock In \emph{Proceedings of the Web Conference 2021}, WWW '21,
  2014–2024. New York, NY, USA: Association for Computing Machinery.
\newblock ISBN 9781450383127.

\bibitem[{Kurtovic(2022)}]{kurtovic2022mwparser}
Kurtovic, B. 2022.
\newblock mwparserfromhell.
\newblock \url{https://github.com/earwig/mwparserfromhell}.
\newblock Accessed: 2022-10-14.

\bibitem[{Li, Lu, and Yin(2022)}]{li2022towards}
Li, Z.; Lu, Z.; and Yin, M. 2022.
\newblock Towards Better Detection of Biased Language with Scarce, Noisy, and
  Biased Annotations.
\newblock In \emph{Proceedings of the 2022 AAAI/ACM Conference on AI, Ethics,
  and Society}, 411--423.

\bibitem[{Lim et~al.(2020)Lim, Jatowt, F{\"a}rber, and
  Yoshikawa}]{lim-etal-2020-annotating}
Lim, S.; Jatowt, A.; F{\"a}rber, M.; and Yoshikawa, M. 2020.
\newblock Annotating and Analyzing Biased Sentences in News Articles using
  Crowdsourcing.
\newblock In \emph{Proceedings of the Twelfth Language Resources and Evaluation
  Conference}, 1478--1484. Marseille, France: European Language Resources
  Association.
\newblock ISBN 979-10-95546-34-4.

\bibitem[{Matei and Dobrescu(2011)}]{matei2011wikipedia}
Matei, S.~A.; and Dobrescu, C. 2011.
\newblock Wikipedia's “neutral point of view”: Settling conflict through
  ambiguity.
\newblock \emph{The Information Society}, 27(1): 40--51.

\bibitem[{Maystre, Kristof, and Grossglauser(2019)}]{maystre2019pairwise}
Maystre, L.; Kristof, V.; and Grossglauser, M. 2019.
\newblock Pairwise Comparisons with Flexible Time-Dynamics.
\newblock In \emph{Proceedings of the 25th ACM SIGKDD International Conference
  on Knowledge Discovery \& Data Mining}, KDD '19, 1236–1246. New York, NY,
  USA: Association for Computing Machinery.
\newblock ISBN 9781450362016.

\bibitem[{Pavalanathan, Han, and Eisenstein(2018)}]{pavalanathan2018mind}
Pavalanathan, U.; Han, X.; and Eisenstein, J. 2018.
\newblock Mind Your POV: Convergence of Articles and Editors Towards
  Wikipedia's Neutrality Norm.
\newblock \emph{Proceedings of the ACM on Human-Computer Interaction}, 2(CSCW):
  1--23.

\bibitem[{Pryzant et~al.(2020)Pryzant, Diehl~Martinez, Dass, Kurohashi,
  Jurafsky, and Yang}]{pryzant2020wnc}
Pryzant, R.; Diehl~Martinez, R.; Dass, N.; Kurohashi, S.; Jurafsky, D.; and
  Yang, D. 2020.
\newblock Automatically Neutralizing Subjective Bias in Text.
\newblock \emph{Proceedings of the AAAI Conference on Artificial Intelligence},
  34(01): 480--489.

\bibitem[{Spinde et~al.(2021)Spinde, Rudnitckaia, Mitrović, Hamborg,
  Granitzer, Gipp, and Donnay}]{SPINDE2021102505}
Spinde, T.; Rudnitckaia, L.; Mitrović, J.; Hamborg, F.; Granitzer, M.; Gipp,
  B.; and Donnay, K. 2021.
\newblock Automated identification of bias inducing words in news articles
  using linguistic and context-oriented features.
\newblock \emph{Information Processing \& Management}, 58(3): 102505.

\bibitem[{Vafa, Naidu, and Blei(2020)}]{vafa-etal-2020-text}
Vafa, K.; Naidu, S.; and Blei, D. 2020.
\newblock Text-Based Ideal Points.
\newblock In \emph{Proceedings of the 58th Annual Meeting of the Association
  for Computational Linguistics}, 5345--5357. Online: Association for
  Computational Linguistics.

\bibitem[{Wikipedia(2022{\natexlab{a}})}]{wiki2022clist}
Wikipedia. 2022{\natexlab{a}}.
\newblock Wikipedia:List of controversial issues.
\newblock
  \url{https://en.wikipedia.org/wiki/Wikipedia:List_of_controversial_issues}.
\newblock Accessed: 2022-08-01.

\bibitem[{Wikipedia(2022{\natexlab{b}})}]{wiki2022w2w}
Wikipedia. 2022{\natexlab{b}}.
\newblock Wikipedia:Manual of Style\/Words to Watch.
\newblock
  \url{https://en.wikipedia.org/wiki/Wikipedia:Manual_of_Style/Words_to_watch}.
\newblock Accessed: 2022-08-01.

\bibitem[{Wikipedia(2022{\natexlab{c}})}]{wikipedia2022npov}
Wikipedia. 2022{\natexlab{c}}.
\newblock Wikipedia:NPOV.
\newblock \url{https://en.wikipedia.org/wiki/Wikipedia:Neutral_point_of_view}.
\newblock Accessed: 2022-10-14.

\bibitem[{Wikipedia(2023{\natexlab{a}})}]{wikipedia2023ores}
Wikipedia. 2023{\natexlab{a}}.
\newblock ORES.
\newblock \url{https://www.mediawiki.org/wiki/ORES}.
\newblock Accessed: 2023-05-15.

\bibitem[{Wikipedia(2023{\natexlab{b}})}]{wikipedia2023GA}
Wikipedia. 2023{\natexlab{b}}.
\newblock Wikipedia: Good article criteria.
\newblock \url{https://en.wikipedia.org/wiki/Wikipedia:Good_article_criteria}.
\newblock Accessed: 2023-05-15.

\bibitem[{Wong, Redi, and Saez-Trumper(2021)}]{wong2021wiki}
Wong, K.; Redi, M.; and Saez-Trumper, D. 2021.
\newblock Wiki-Reliability: A Large Scale Dataset for Content Reliability on
  Wikipedia.
\newblock In \emph{Proceedings of the 44th International ACM SIGIR Conference
  on Research and Development in Information Retrieval}, 2437--2442. New York,
  NY, USA: Association for Computing Machinery.

\bibitem[{Yardim et~al.(2018)Yardim, Kristof, Maystre, and
  Grossglauser}]{yardim2018wikipedia}
Yardim, A.~B.; Kristof, V.; Maystre, L.; and Grossglauser, M. 2018.
\newblock Can Who-Edits-What Predict Edit Survival?
\newblock In \emph{Proceedings of the 24th ACM SIGKDD International Conference
  on Knowledge Discovery \& Data Mining}, KDD '18, 2604–2613. New York, NY,
  USA: Association for Computing Machinery.
\newblock ISBN 9781450355520.

\bibitem[{Zhong et~al.(2021)Zhong, Yang, Xu, and Yang}]{zhong2021wikibias}
Zhong, Y.; Yang, J.; Xu, W.; and Yang, D. 2021.
\newblock WIKIBIAS: Detecting Multi-Span Subjective Biases in Language.
\newblock In \emph{Findings of the Association for Computational Linguistics:
  EMNLP 2021}, 1799--1814. Punta Cana, Dominican Republic: Association for
  Computational Linguistics.

\end{thebibliography}

\end{document}